\documentclass[runningheads]{llncs}
\usepackage{graphicx}

\usepackage{tikz}
\usepackage{comment}
\usepackage{amsmath,amssymb} 
\usepackage{color}

\usepackage[accsupp]{axessibility}  

\usepackage{booktabs}
\usepackage{array,multirow}
\usepackage{soul}
\usepackage{enumitem}

\newcommand{\jd}[1]{{\color{SeaGreen}#1}}

\usepackage[colorinlistoftodos]{todonotes}
\newcommand{\JD}[1]{{\todo[size=\tiny]{DJ: #1}}}

\newenvironment{tight_itemize}{
\begin{itemize}[leftmargin=15pt,nosep]
  \setlength{\topsep}{0pt}
  \setlength{\itemsep}{0pt}
  \setlength{\parskip}{0pt}
  \setlength{\parsep}{0pt}
}{\end{itemize}}

\usepackage[labelfont=bf]{caption}
\DeclareMathOperator*{\argmin}{argmin}

\begin{document}
\pagestyle{headings}
\mainmatter
\def\ECCVSubNumber{5255}  

\title{Long-HOT: A Modular Hierarchical Approach for Long-Horizon Object Transport} 

\titlerunning{Long-HOT}
%
\author{Sriram Narayanan$^1$ \and 
Dinesh Jayaraman$^2$  \and
Manmohan Chandraker$^{1,3}$
}
\authorrunning{Narayanan et al.}
%

\institute{NEC Labs America \and 
University of Pennsylvania \and 
UC San Diego
}
\maketitle

\begin{abstract}
We address key challenges in long-horizon embodied exploration and navigation by proposing a new object transport task and a novel modular framework for temporally extended navigation. Our first contribution is the design of a novel Long-HOT environment focused on deep exploration and long-horizon planning where the agent is required to efficiently find and pick up target objects to be carried and dropped at a goal location, with load constraints and optional access to a container if it finds one. Further, we propose a modular hierarchical transport policy (HTP) that builds a topological graph of the scene to perform exploration with the help of weighted frontiers. Our hierarchical approach uses a combination of motion planning algorithms to reach point goals within explored locations and object navigation policies for moving towards semantic targets at unknown locations. Experiments on both our proposed Habitat transport task and on MultiOn benchmarks show that our method significantly outperforms baselines and prior works. Further, we validate the effectiveness of our modular approach for long-horizon transport by demonstrating meaningful generalization to much harder transport scenes with training only on simpler versions of the task. 


\end{abstract}

\section{Introduction}
\begin{figure}[!t]
    \centering
    \includegraphics[width=1.0\linewidth, trim={1cm 0.8cm 4.7cm 0.5cm}, clip]{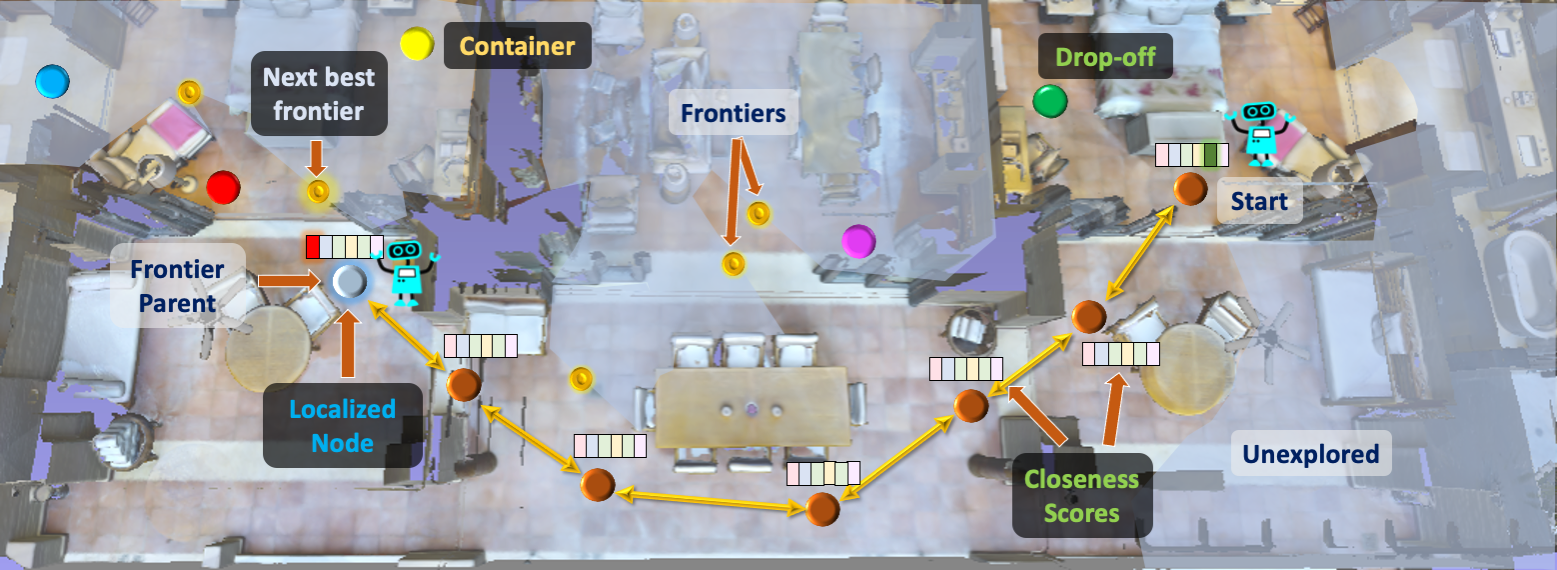}
    \caption{
    An instance of object transport problem and our proposed modular approach that builds a topological map. 
    If the agent decides to explore a bit further it would then detect other objects and a container which it can pick up to transport more efficiently, rather than dropping just one object at the goal. 
    }
    \label{fig:teaser}
\end{figure}

A robot tasked with finding an object in a large environment or executing a complex maneuver must reason over a long horizon. Existing end-to-end reinforcement learning (RL) approaches often suffer in long-horizon embodied navigation tasks due to a combination of challenges: (a) inability to provide exploration guarantees when the point or object of interest is not visible, (b) difficulty in backtracking previously seen locations and (c) difficulty in planning over long horizons. In this work, we address these issues by proposing a novel long-horizon embodied transport task, as well as modular hierarchical methods for embodied transport and navigation.

Our proposed long-horizon object transport task, Long-HOT, is designed to study modular approaches in the Habitat environment \cite{habitat_iccv2019}. It requires an embodied agent to pick up objects placed at unknown locations in a large environment and drop them at a known goal location, while satisfying load constraints, which may be relaxed by picking up a special container object (Fig.~\ref{fig:teaser}). While tasks like MultiON for sequential navigation also benefit from long-range planning \cite{wani2020multion}, the proposed transport task requires more complex decision-making, such as order of pick-up and exploration-exploitation trade-off with respect to searching for the container. We abstract away the physical reasoning for pickup and drop actions, since unlike TDW-Transport \cite{threedworld_transport}, our focus is on deeper exploration and long-horizon planning to find and transport objects in large scenes.

We argue that modularity is a crucial choice for tackling the above challenge, whereby navigation and interaction policies can be decoupled through temporal and state abstractions that significantly reduce training cost and enhance semantic interpretability compared to end-to-end approaches. This is distinct from existing hierarchical methods for subgoal generation \cite{krantz2021waypoint,xia2021relmogen} in long horizon tasks, where the expressivity of subgoals is largely limited to goal reaching for embodied navigation and which still face scalability challenges when the task requires long trajectory demonstrations. 

Our modular approach for long-horizon embodied tasks constitutes a topological graph based exploration framework and atomic policies to execute individual sub-tasks (Fig.~\ref{fig:teaser}). The higher level planner is a finite state machine that decides on the next sub-routine to execute from one of \texttt{\small \{Explore, Pickup, Drop\}} actions. The topological map representation consists of nodes connected in the form of a graph that serves to infuse geometric and odometry information to aid deeper exploration and facilitate backtracking. Unlike methods that utilize 360-degree panoramic images as input \cite{chaplot2020neural,VGM,krantz2021waypoint}, we divide every node to aggregate representations from several directions in its vicinity. The representation within a specific node and direction consists of latent features $\mathcal{F}_A$ from a pre-trained encoder, an exploration score $\mathcal{F}_E$ that captures the likelihood of the agent finding an object if it explores a frontier in that direction and object closeness score $\mathcal{F}_O$ that indicates distance to the object within the agent's field of view. We also propose a novel weighted improvement of frontier exploration \cite{frontier} using the predicted exploration scores.

Unlike methods \cite{xia2021relmogen,krantz2021waypoint,VGM,habitat2o} that completely rely on either motion planning algorithms \cite{planningalgo,roboticsbook} or use pure RL for low-level actions \cite{VGM,wani2020multion}, our approach uses the best of both worlds with motion planning for point goals within explored regions and RL policies to travel the last-mile towards semantic targets at unknown locations. Indeed, on both Long-HOT and MultiON , we show that our proposed modular hierarchical approach performs significantly better, especially on longer horizons, compared to agents operating without map or other hierarchical approaches that sample navigation subgoals for task completion. Moreover, it realizes a key benefit of modularity, namely, adaptability to harder settings when trained on a simpler version of the task.

In summary, our contributions are the following:
\begin{tight_itemize}
    \item A novel object transport task, Long-HOT, for evaluating embodied methods in complex long-horizon settings.
    \item A modular transport framework that builds a topological graph and explores an unknown environment using weighted frontiers.
    \item Strong empirical performance on object transport and navigation tasks, with adaptability to longer horizons . 
\end{tight_itemize}

\section{Related Work}
\paragraph{\bf Embodied intelligence.} 

The community has developed several simulation environments \cite{habitat_iccv2019,ai2thor,shen2021igibson,threedworld_transport,chen2020soundspaces,xiazamirhe2018gibsonenv,RoboTHOR} and associated tasks to study embodied agents in tasks like object goal navigation \cite{batra2020objectnav,chaplot2020object,wortsman2019learning,yang2018visual,wani2020multion}, point goal navigation \cite{anderson2018evaluation,habitat_iccv2019,wijmans2020ddppo,ramakrishnan2020occupancy}, rearrangement \cite{threedworld_transport,shridhar2020alfred,Weihs_2021_CVPR,habitat2o}, instruction following \cite{shridhar2020alfred,anderson2018visionandlanguage} and several others in this regard.
While there are handful of previous works designed for navigation \cite{batra2020objectnav,chaplot2020object,wani2020multion} or rearrangement \cite{threedworld_transport,Weihs_2021_CVPR}, they do not extensively stress tests methods with increasing task complexities. We find the typically used flat policy architectures \cite{wani2020multion} in embodied AI tasks fail completely when executing over longer horizons. Hence, we propose a new benchmark called Long-HOT that has potential to serve as a testbed for, and accelerate the development of novel architectures for planning, exploration, and reasoning over long spatial and temporal horizons. 


Our task builds on previous transport tasks defined in embodied intelligence \cite{Weihs_2021_CVPR,habitat2o,threedworld_transport,wani2020multion} but differs in ways that it requires deeper exploration and long horizon planning. While previous work like \cite{Weihs_2021_CVPR} focus on identifying state changes using visual inputs to perform rearrangement or \cite{habitat2o} use geometrically specified goals in single apartment environments these works operate in minimal exploration scenarios where the focus is shifted more towards perception or interaction with objects. 
Our task is closest to \cite{threedworld_transport}, while \cite{threedworld_transport} focuses on performing transport including physics based simulations, we abstract our interactions and focus more on complex long-horizon planning.
Our work extends \cite{wani2020multion} but rather than focusing on navigation in a predefined sequential fashion, our task requires more complex decision making to determine the order of picking and decide whether to perform a greedy transport if it sees the goal or to explore more in hopes of finding the container for efficient transport.




    

\paragraph{\bf Modular-Hierarchical Frameworks.} 
Solutions to long-horizon tasks typically involved hierarchical\cite{NIPS1997_hierarchy,subgoal_discovery,Sutton:1999,bacon2016optioncritic} policies in reinforcement learning. They provide better exploration behavior through long-term commitment towards a particular subtask. \cite{xia2021relmogen,krantz2021waypoint} present one such approach where they sample navigation subgoals to be executed by the low-level controller. While these methods can temporally abstract navigation to an extent we find their performance to drop significantly in longer horizon settings. In HTP we show that modularity enables generalization while only training on the simpler versions of the task. 


Closer to our work are modular approaches\cite{NMCdas2019,chaplot2020neural,krantz2021waypoint,xia2021relmogen} that provide an intuitive way to divide complex long horizon tasks as a combination of individual sub-tasks that can be solved using existing approaches. Das et al. \cite{NMCdas2019} present a  modular approach to solve embodied question answering \cite{das2017embodied} through a combination of several navigation policies each for finding an object, to find a room or to exit one. 
This can blow-up with number of sub-routines required to navigate across a building or inability of the agent to find a room of particular type in large environments. Rather than navigating to individual rooms our method proposes a weighted frontier technique that provides exploration guarantees.
Chaplot et al. \cite{chaplot2020neural} propose a method for image goal navigation by generating a topological map of the scene using 360-degree panoramic images. Our approach operates on perspective images and divides a node representation into segments across different directions.

Our work also closely relates to task and motion planning (TAMP) literature\cite{garrett2020integratedtamp,ffrob} where the closest work in this domain is \cite{yamada2020motion} which proposes a MP augmented RL approach in manipulation settings where they realize large displacements of a manipulator through a MP. 
While \cite{yamada2020motion} tackles a simple manipulation domain for 2D block pushing where target objects are fully observable, we tackle a more complex navigation setting and propose to use a combination of MP and object navigation policies where a motion planner first moves to a region with high likelihood of object presence and gives control to the navigation policy that takes it closer to the goal object.
\section{Habitat Transport Task}
We propose a novel transport task for embodied agents that simulates object search, interaction, and transport in large indoor environments. A robot assistant might be expected to perform such tasks in a warehouse or a hospital.
In each episode, the agent must transport $K$ target objects to a specified object goal location in a large partially observed Habitat~\cite{habitat_iccv2019} 3D indoor environment with many rooms and obstacles.
The environment also contains a special ``container'' object (in yellow) that can be used to transport other objects, another special goal object (in green) whose position is the goal location. In our setting, all objects are cylinders of various colors, placed into the environment.
Unlike previously studied tasks such as~\cite{habitat2o}, the agent needs to explore the environment to find all objects, and does not have access to their geometric coordinates. 
At each step, the agent can turn by angle $\alpha=30^{\circ}$ to the left or right, move forward by $x=0.25m$, or execute object ``Pickup'' or ``Drop'' actions.  


The agent has access to standard egocentric perspective RGB and depth views. Fig.~\ref{fig:pipeline} (left) shows an example of the agent's view of a scene with a prominent red object. Aside from this, the agent has access to odometry ($P_{\phi xy}$),
as well as the hand state $O_h$ and goal state $O_g$, which indicate if a target/container object is either held by the agent or already at the goal respectively. 
Following \cite{wani2020multion,Weihs_2021_CVPR}, if the agent is within $R = 1.5m$ of any pick-able object and a Pickup action is called, the closest object is removed from the scene and the hand state $O_h$ is updated to include it. For the Drop action, any objects in the agent's hand are dropped near the agent's location. If the goal is within distance $R$, the goal state is updated to include the object.
The agent can hold limited items in its hands at once, and is therefore constrained to carry at most two objects at a time unless it picks up the container, in which case any number of objects may be carried. Picking up the container requires the agent's hands to be empty. Each episode runs for a maximum of $T = 2500$ timesteps.


This transport task naturally entails additional complexity compared to previously proposed navigation settings, and has properties that are not emphasized in previous benchmark tasks for embodied agents~\cite{wani2020multion,habitat2o,habitat_iccv2019,xiazamirhe2018gibsonenv}. It includes multiple task phases (searching for, navigating to, and interacting with objects), reasoning about the environment at various scales (such as coarse room connectivity charts over the explored map for planning long trajectories, and fine-grained local occupancy maps for object interaction), accounting for carrying capacity constraints. It also involves dynamically selecting among various sub-task sequences under uncertainty: for example, having found an object far from the goal, should an agent immediately return to drop it off at the goal, or should it look for another object before returning for efficiency?

\section{Hierarchical Transport Policy (HTP)}
\begin{figure}[!!t]
    \centering
    \includegraphics[width=1.0\textwidth, trim={0 0 0 0},clip]{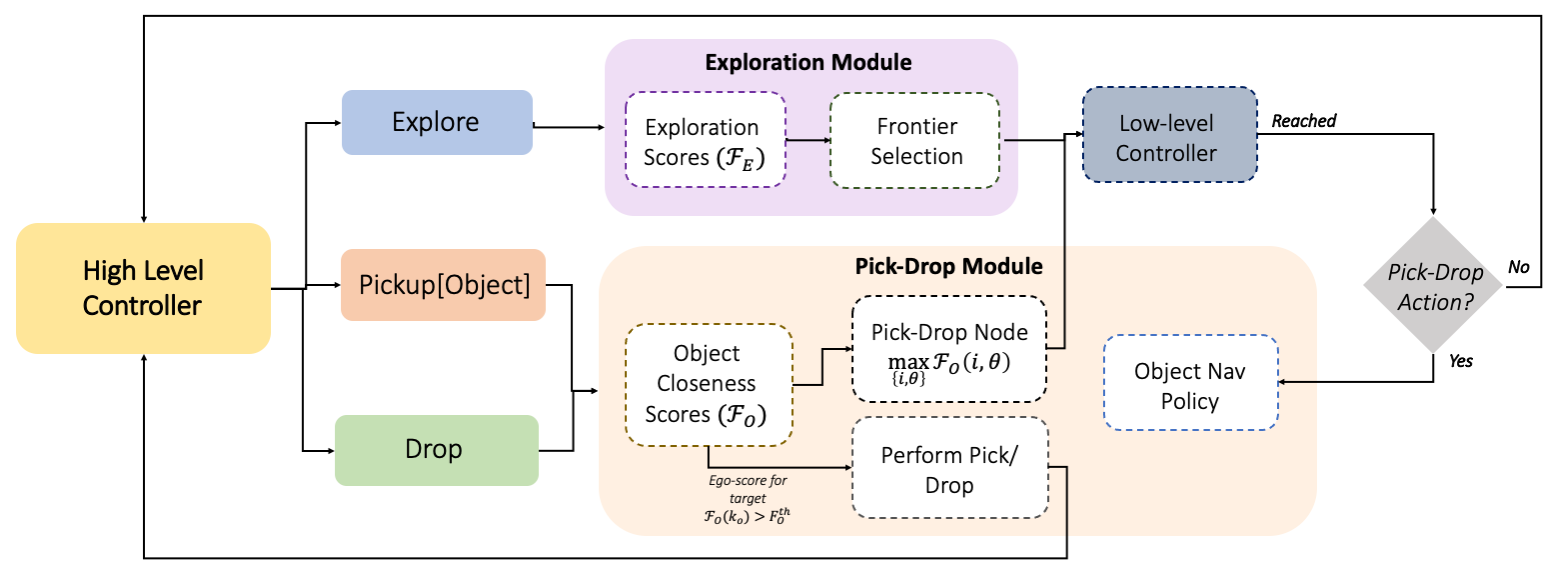}
    \caption{Overall architecture of our proposed modular hierarchical approach. 
    Our framework consists of three main modules a) a high level controller, b) exploration module  and c) a pick-drop module. Each of these are made of components like exploration and closeness score predictor, object-nav policy and low-level controller. We provide details for each individual component in Sec. \ref{ssec:components} and explain the overall framework in Sec. \ref{ssec:method}.
    }
    \label{fig:framework}
\end{figure}

We now describe a modular policy (Fig. \ref{fig:framework}) that builds a topological map of the environment and operates at different levels of temporal abstraction. Our framework consists of three modules: {\it a high-level controller}, {\it an exploration module} and {\it a pick-drop module}. The {\it high-level controller} decides on the next high-level action to execute from one of $\mathcal{A}_H =$\texttt{\small \{Explore, Pickup[Object], Drop\}} actions. The appropriate sub-task module then takes over to perform the given high-level action. At any point, if the high-level controller decides to execute a different sub-task, the current execution is interrupted and control is passed to the module executing the next one. The modules in our framework are made of several functions which we describe briefly in Sec. \ref{ssec:components} and then provide details on how those components are connected to our overall framework.

\subsection{HTP Model Components}\label{ssec:components}
\begin{figure*}[!!t]
    \centering
    \includegraphics[width=1.0\textwidth, trim={0 0 3cm 0}, clip]{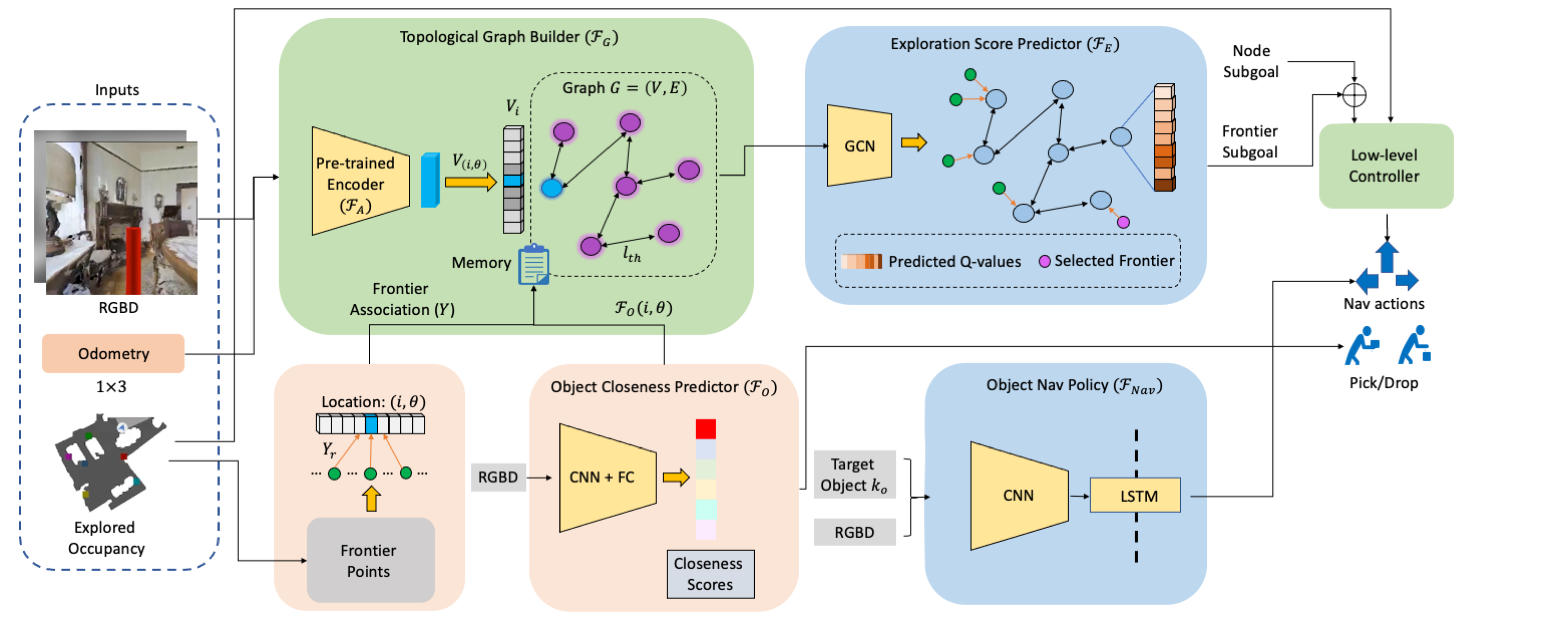}
    \caption{Shows an overview of the model components used in our overall framework along with their connectivity. It takes egocentric RGBD information along with odometry and explored occupancy map as input and provides low-level actions executed by the agent. 
    At every timestep the framework builds a topological map (Sec. \ref{ssec:graphupdate}), predicts object closeness scores (Sec. \ref{sec:score_pred}), sample frontiers based on exploration scores (Sec. \ref{sec:score_pred}) and execute navigation actions through a object-nav policy (Sec. \ref{ssec:objectnav}) or low-level controller (Sec. \ref{ssec:lc}).
    }
    \label{fig:pipeline}
\end{figure*}

An overview of our model components along with their connections in the HTP framework is shown in Fig. \ref{fig:pipeline}. Our sub-task modules consist of the following components: a) a topological graph builder, b) exploration score predictor c) object-closeness predictor, d) object navigation policy and e) a low-level controller. We now describe each of them in detail. 

\subsubsection{Topological Graph Builder}\label{ssec:graphupdate}
This function is responsible for creating a topological map of the environment as a graph ${G} = ({V}, {E})$, where $V \in \mathbb{R}^{N_t\times f_d}$ and $E$ represent spatial nodes and connecting edges respectively. Here, $N_t$ represents number of nodes at timestep $t$ and $f_d$ represents the length of node features. Node features for a node $V_i \in {V}$ consist of concatenated ``node-direction'' features $V_{i, \theta}$ corresponding to $D=12$ directions $\theta \in \{1,...,D\}$ spanning 360 degrees, centered on the node. 
These node-direction features $V_{i, \theta}$ in turn are computed by encoding perspective RGB images through an encoder $\mathcal{F}_A$, pretrained in an autoencoder (details in supplementary).




At every timestep $t$, the graph builder updates the map as follows. 
It takes the pose information ($P_{\phi xy}$)
and encoded egocentric image features from $\mathcal{F}_A$ as input. 
The agent's location $P_{xy}$ is mapped to the nearest existing graph node $V_i$, and its heading angle $P_{\phi}$ is mapped onto the nearest node-direction $\theta$. Then that corresponding node-direction representation 
$V_{i,\theta}$ is updated to the image feature vector. 
We add a new node $V_{i+1}$ if the agent is not within a distance threshold $l_{th} = 2m$ of existing nodes, and store the corresponding node coordinate $P_{xy}$. 
When the agent transitions between two nodes in the graph, the graph is updated to add an edge between them.
Similar to \cite{VGM}, we also keep track of the last visited time-step for every node to provide additional context for downstream processing. 

\subsubsection{Exploration and Object Closenes Scores}\label{sec:score_pred}\label{ssec:scorepred}
At every node-direction, indexed by $(i, \theta)$, aside from the feature vector above, we store two score predictions: (a) an {\it exploration score} for frontiers\cite{frontier} that predicts the likelihood of finding an object by exploring that frontier and (b) an {\it object closeness score} that indicates distance to various objects in the current field of view. 

\vspace{-0.2cm}
\paragraph{Exploration Score Predictor ($\mathcal{F}_E$):}\label{para:exp_score_pred} 
We explore two variants of this function: (1) a Q learning based graph convolutional network (GCN)\cite{kipf2017semi} that reasons for frontiers over the entire topological graph and (2) a convolutional network (CNN) predicts frontier scores on a per-frame basis.

\vspace{0.2cm}
\noindent{\it GCN Exploration Score:} This function operates on (1) 
the current topological map
$G = (V,E)$ with associated features as computed above, and (2) a binary mask $M^{N_t \times \theta_d}$ indicating the availability of a frontier at every node-direction, computed using the method described in Sec. \ref{ssec:expmodule}.
Provided these inputs, we train a graph convolutional network (GCN)~\cite{kipf2017semi} to produce reinforcement learning Q-values for each node-direction, representing future rewards for finding objects after visiting each frontier associated with that node-direction. 
The object-finding reward function $r_t^e$ at every timestep $t$ is:
\begin{equation}
    r_t^e = \mathbb{I}_{success} \cdot r^e_{success} +  r^e_{slack} + \sum_o \mathbb{I}^o_{found} \cdot r^e_{found},
\end{equation}
where $\mathbb{I}^o_{found}$ is the indicator if object $o$ was found at timestep $t$, $r^e_{found}$ is the reward for finding a new object, $r^e_{slack}$ is the time penalty for every step that encourages finding the objects faster, $\mathbb{I}_{success}$ is an indicator if \textit{all} objects were found and $r^e_{success}$ is the associated success bonus. We consider the object to be found if the object is in the agent's field of view with distance less than maximum pre-defined distance. 
Our GCN architecture involves three layers of graph convolution layers and a fully connected layer.


\vspace{0.2cm}
\noindent{\it CNN Exploration Score:} This variant of the exploration score is computed directly from the agent's current RGBD view. 
Given this view, a CNN predicts three exploration scores:
each score represents the chances of finding an object if the agent explores the farthest frontier available within a corresponding range of angles centered on the current agent heading: $-45^\circ$ to $-15^\circ$, $-15^\circ$ to $+15^\circ$, and $+15^\circ$ to $+45^\circ$ respectively for the three scores. This CNN is trained with labels set to $\max(\max_o(d_{a,o} - d_{f,o})/5, 0)$ where $d_{a,o}, d_{f,o}$ represent geodesic distance to object $o$ from the agent and frontier respectively. If a frontier is not available then the score is set to 0. These three scores are then stored respectively to three consecutive node-directions $\theta-1, \theta, \theta+1$, centered on the current direction $\theta$, and at the current node $i$.  
\vspace{-0.2cm}
\paragraph{Object Closeness Predictor ($\mathcal{F}_O$):} This CNN maps the current RGBD observation $I$ to a ``closeness score'' for every object. It is trained with supervised learning to predict target closeness labels for each object, which are set to $\max(1-d/5, 0)$ where $d$ is the true distance to the object in m. So, objects farther than $5m$ away (or invisible) have labels $0$, and very close objects have labels $\approx 1$. Each node-direction has an associated closeness score for each object. 

\subsubsection{Object Navigation Policy}\label{ssec:objectnav}
Next, we discuss our object navigation policy. Given the RGBD observation $I$ and a one-hot encoding $k_o$ of a target object, the policy must select navigation actions from \texttt{\small \{FORWARD, TURN-LEFT, TURN-RIGHT\}} that take it closer towards the target. This policy is trained with the following reward $r^n_t$ \cite{wani2020multion} at each timestep $t$:
\begin{equation}
r^n_t = \mathbb{I}_{[reached-obj]} \cdot r^n_{obj} + r^n_{slack} + r^n_{d2o} + r^n_{collision},
\end{equation}
where $r^n_{obj}$ is the success reward if it reaches closer than a threshold distance $d_{th}$ with the target object, $r_{slack}$ is a constant time penalty for every step, $r^n_{d2o} = (d_{t-1} - d_t)$ is the decrease in geodesic distance with the target object and $r^n_{collision}$ is the penalty for collision with the environment.  

We train this policy using the proximal policy optimization (PPO) \cite{ppo} reinforcement learning algorithm, for approximately 40M iterations using 24 simulator instances. We use mini-batch size of 4 and perform 2 epochs in each PPO update. We use other hyper-parameters similar to \cite{wani2020multion}.

\subsubsection{Low-Level Navigation Controller}\label{ssec:lc}
Our final module is a low-level controller that takes a goal location (from within the explored regions)
to be reached as input. 
It then plans a path towards the specified goal location using the classical A*\cite{astar} planning algorithm using a pre-built occupancy map.

\subsection{HTP Control Flow}\label{ssec:method} 
We are now ready to describe how  HTP manages the flow of control between these components to perform long-horizon transport tasks.
Note that while we describe the HTP algorithm for object transport, we show in Sec.~\ref{ssec:quant_compare} that HTP also works for other embodied navigation tasks.



\subsubsection{High-Level Controller}\label{ssec:highlevel}
The high-level controller ($\pi^H$) is a finite state machine. Based on object closeness scores $\mathcal{F}_O$ (Sec. \ref{sec:score_pred}), hand state $O_h$, and goal state $O_g$,
it selects one subtask from among $\mathcal{A}_H =$\texttt{\small \{Explore, Pickup[Object], Drop\}}. 
At timestep $t$, if the next high level action predicted by the controller is different from the current sub-task that is being executed, the controller interrupts the execution, and agent performs the updated high-level action. For example, during exploration if the agent finds an object with closeness score higher than a some threshold it then switches control from exploration to picking the object if the hand is not full or if it holds a container. 



\subsubsection{Weighted Frontier  Exploration}\label{ssec:expmodule}

If $\pi^H$ selects the $<$\texttt{\small Explore}$>$ sub-task, the exploration module is executed. For exploration, we introduce a weighted frontier technique based on the predicted exploration score function $\mathcal{F}_E$ (Sec \ref{sec:score_pred}). For every timestep $t$, we calculate the set of frontiers ${\bf S}$ over the explored and unexplored regions using occupancy information \cite{frontier}.
When a new frontier is identified, we assign a parent node-direction $Y_r = (i,\theta^n)_r$ for the $r^{th}$ frontier, where $(i,\theta)$ is the current localized node-direction and $\theta^n$ is calculated based on the angle made by the frontier with the agent. Here the agent's field of view is $90^\circ$, so $\theta^n$ for the newly found frontiers can assume one of $\{\theta-1,\theta,\theta+1\}$ directions.
For all existing frontiers from timestep $t-1$, we copy the same parent node-direction from the previous timestep. Finally, we calculate a representative frontier $ S^{(i,\theta)}$ for node-direction $(i,\theta)$, as: $S^{(i,\theta)} = \{s_k: \argmin_k \|s_k - X_c\|\ \forall\ Y_k = (i,\theta)\}$ where $s_k \in {\bf S}$ and $X_c$ is the center of frontiers associated with $Y_k=(i,\theta)$. 


At each timestep during its execution, the exploration module selects a node-direction $(i,\theta)$ from the topological graph $G$ that has the highest exploration score $\mathcal{F}_E$. Its corresponding frontier $S^{(i,\theta)}$ is then set as the goal location for the agent's low-level controller, which begins to move towards this goal.  
The highest-score goal frontier is recomputed at every timestep, and may switch as new views are observed during exploration. 


\subsubsection{Pick-Drop Module}\label{ssec:pickdrop}
This module performs the pick or drop actions in the object transport task when the controller $\pi^H$ selects an action $a_H \in$\ \{\texttt{\small Pickup[Object], DropAtGoal}\}. 
When called, this module first selects a node $(i,\theta)$ from graph $G$ with the highest object closeness score $\mathcal{F}_O$ for the target object. 
If the agent is not already in the selected $i^{th}$ node, then its location $P_{xy}(i)$ is set as the goal for the low-level controller. Once the agent is localized to the $i^{th}$ node, it orients in the direction of $\theta$. At this point, control is passed to the Object Navigation policy, targeting the object selected by $\pi_H$. 
The module then selects the pickup or drop action whenever the object closeness score $\mathcal{F}_O$ for the target object, based on the current view, exceeds a threshold. 
The sub-task is successful when the hand state or goal state is changed accordingly and the controller $\pi^H$ predicts the next high level action to execute.  We execute this module till it performs the pick/drop or for a maximum of $T_p$ steps after which the control is given back to the high-level controller $\pi^H$.

\section{Experiments}
\begin{figure*}[!!t]
    \centering
    \includegraphics[width=1.0\textwidth, trim={0 0 0 1cm}, clip]{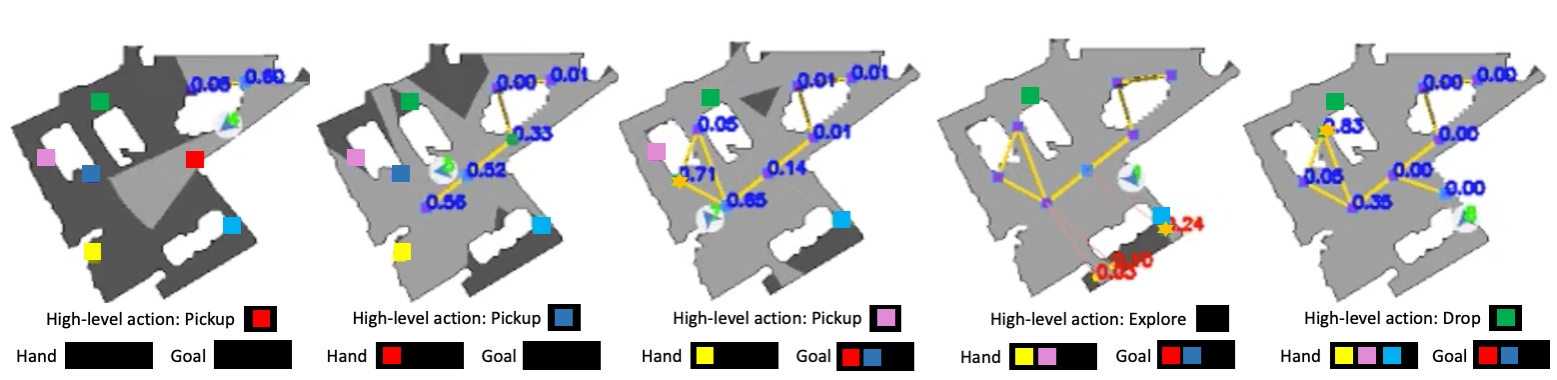}
    \caption{
    An example execution of our proposed Hierarchical policy (HTP) for object transport. Yellow lines indicate edges and their end point indicate node locations. The goal object is green, container is yellow and target objects take other colors. A blue value near the node indicates the closeness score for the object to be picked up specified by the high level action. Yellow circles indicate frontiers and red value indicate their associated exploration scores. A star marker with orange color indicates the current subgoal location for A*. Please note high closeness score for the node close to the object type to interact with specified by the high level action.
    }
    \label{fig:example_results}
\end{figure*}

We evaluate all approaches on two tasks set in photo-realistic Matterport3D scenes (MP3d)\cite{Matterport3D} in Habitat~\cite{habitat_iccv2019}: Long-HOT transport, and Multi-ON~\cite{wani2020multion} object navigation. 


\vspace{0.4cm}
\noindent \textbf{Long-HOT:}
We split MP3D\cite{Matterport3D} scenes into disjoint train, validation, and test scenes~\cite{anderson2018evaluation} each with 61/14/15 scenes respectively. We generate 10,000 training task configurations among the training scenes, and 3000 each of validation and test configurations. 
%
%
Each task configuration consists of a specific configuration of objects, container, goal location, and agent starting location and pose. 
First, we sample a goal $(x,y)$ location in the map, then sample the four object locations. These object locations (1) lie in a specified range of distances from the goal (``goal-range''), (2) at a specified minimum distance  (``obj-dist-min'') away from other objects, and at (3) within a specified maximum distance (``obj-dist-max'') from at least one other object. Next, container and agent starting locations are also sampled to lie within the same goal-range as objects. All distances are geodesic. These settings permit modulating complexity: for example, large goal distances lead to harder tasks that are more exploration-intensive and need a longer task horizon. 
Table \ref{tab:dataset} shows settings for different task levels used in our experiments. We train all methods on \emph{default}-level tasks on 61 scenes. 
After training, we first evaluate them on 15 disjoint test scenes in the \textit{default} setting and call this as ``Standard Long-HOT Task''.
We then perform a more focused ``Large Long-HOT Task'' evaluation on \textit{large} scenes that have at least one dimension $>40m$ and sample \emph{default}, \emph{hard} and \emph{harder} level tasks from Table \ref{tab:dataset}.


\begin{table}[h]
\centering
\caption{Specification parameters for our task levels}

\begin{tabular}{lcccc}
\toprule
Level & goal-range(m) & obj-dist-min/max(m)\\
\midrule
\textit{default} & (2, 15) & 2/10\\
\textit{hard} & (5, 20) & 5/20\\
\textit{harder} & (5, 30) & 5/30\\
\bottomrule


\end{tabular}
\label{tab:dataset}
\end{table}

\vspace{0.4cm}
\noindent \textbf{MultiOn Dataset:}
MultiOn\cite{wani2020multion} is a sequential multi-object navigation task where the agent is required to visit objects in a predefined sequence.
We adapt our proposed transport policy to test it on the challenging MultiOn (3-On) task using the object goal vector as input to the  high-level controller (more details in supplementary). 

\begin{table*}[]
\centering
\caption{\textbf{Standard Long-HOT Task:} Comparison of metrics for the proposed HTP along with the baselines in standard transport settings. }
\resizebox{1.0\textwidth}{!}{
\begin{tabular}{lcccccc}
\toprule
Model                                     & \%Success$\uparrow$ & \%Progress$\uparrow$  & SPL$\uparrow$ & PPL$\uparrow$  & Episode energy$\downarrow$ & \%Picked$\uparrow$ \\
\midrule
OracleMap (Occ)                        & 56      & 68         & 34   & 37   & 0.34      & 69       \\
OracleMap (Occ+Obj)              & 92      & 95         & 74 & 75 & 0.05      & 96       \\
OracleMap-Waypoints & 85      & 90         & 51 & 52 & 0.10      & 93       \\
\midrule
NoMap                                   & 43      & 63         & 26 & 32 & 0.41      & 67       \\
ProjNeuralMap\cite{wani2020multion} & 43 & 60 & 24 & 29 & 0.42 & 60\\
\midrule
HTP - NearestFrontier & 52 & 66 & 24 & 29 & 0.38 & 75\\
HTP - CNN & 56      & {\bf 72}         & 25 & 31   & {\bf 0.32}      & {\bf 79}\\
HTP - GCN & {\bf 59} & 70 & {\bf 28} & {\bf 33} & 0.33 & 77\\
\bottomrule
\end{tabular}
}
\label{tab:standard}
\end{table*}

\vspace{0.4cm}
\noindent \textbf{Baselines:} We compare our proposed transport policy with several baseline methods and ablations. 
\begin{itemize}[leftmargin=*]
\item \textbf{NoMap:} This baseline policy, trained using PPO\cite{ppo}, maps RGBD image $I$, hand state $O_h$ and goal state $O_g$ directly to low-level robot actions.
\item \textbf{OracleMap:} This method improves NoMap by assuming additional access to a ground truth 
2D occupancy map of the $10m\times 10m$ area centered on the agent in the overhead view. Similar to \cite{wani2020multion}, we evaluate two versions of OracleMap: with occupancy alone (``Occ''), and with extra annotated true locations of the task-relevant objects and container (``Occ+Obj''). 
\item \textbf{OracleMap-Waypoints:} This baseline represents the a popular hierarchical approach in embodied navigation~\cite{krantz2021waypoint,xia2021relmogen}: setting navigation waypoints for a motion planner, such as A\text{*}~\cite{astar}. It trains an RL policy to select discretized $(x, y)$ waypoints on the map. We provide access to OracleMap (Occ+Obj) for this baseline. See supplementary for details.

\item \textbf{MultiOn Baselines:} For Long-HOT object transport, we adapt the ProjNeuralMap baseline from \cite{wani2020multion} that projects perspective features in top view to our task. To this, we add additional hand-state and goal state embeddings instead of object goal embeddings as in MultiOn.
For MultiOn, we compare against the authors' baselines~\cite{wani2020multion}, as well as the best-performing methods from the public leaderboard.

\item \textbf{Exploration Ablations:} We study three variants of our method with different exploration strategies: NearestFrontier, CNN, GCN. HTP-NearestFrontier uses vanilla frontier exploration\cite{frontier} and picks the closest frontier from the agent location as the next exploration subgoal. HTP-CNN and HTP-GCN use our proposed CNN and GCN-based exploration scores for weighted frontier exploration, explained in Sec~\ref{sec:score_pred}.


\end{itemize}


\subsection{Metrics}
We use standard evaluation metrics following previous works \cite{Weihs_2021_CVPR,wani2020multion,gupta2019cognitive,jain2020cordial,chen2020soundspaces,anderson2018evaluation,habitat_iccv2019} and adapt a few other metrics relevant to our task setting. 
\vspace{-0.3cm}
\paragraph{\bf \%Success:} It measures the percentage of successful episodes across the test set. An episode is successful if the agent moves all $K$ objects to the goal location. 

\vspace{-0.3cm}
\paragraph{\bf \%Progress:} It measures the percentage of target objects successfully transported to the goal location.

\vspace{-0.3cm}
\paragraph{\bf SPL \& PPL:} SPL is Success weighted-by Path Length, and PPL is Progress weighted by Path Length. Since there multiple ways in which one can complete this task we substitute optimal path length in SPL and PPL calculations with a reference path length $G_{ref}$ (details in supplemetary). Any execution with path length $G_{pl} \leq G_{ref}$ weights the success and progress values by $1.0$. Hence  $\texttt{SPL} = 1_{\texttt{success}}\times \min(G_{pl}/G_{ref},1.0)$ and $\texttt{PPL} = \texttt{Progress}\times \min(G_{pl}/G_{ref},1.0)$. 

\vspace{-0.3cm}
\paragraph{\bf Episode Energy:} We adapt a similar metric from \cite{Weihs_2021_CVPR} to our task setting. 
It measures the amount of remaining energy to complete the episode and gives partial credit if the agent successfully moves the object closer to goal. It is defined as $E = \sum_{k=1}^K d_{g2t^k}/ \sum_{k=1}^K D_{g2t^k}$ where numerator and denominator represent sum of geodesic distance of target objects to goal location at the ending and starting of the episode respectively. 

\vspace{-0.3cm}
\paragraph{\bf \% Picked:} This metric measures the percentage of target objects that are successfully picked.
\begin{table}[t!]
\caption{\texttt{Left:}\textbf{Large Long-HOT Task.} Shows generalizability of methods to more difficult settings within the same task. 
The methods were tested on 26 habitat scenes with at least one dimension $>40m$. All methods were trained on dataset generated with \textit{default} task configuration (max. radius 15m) and tested on \textit{hard} (max. radius 20m) and \textit{harder} (max. radius 30m) transport settings. \texttt{Right:} MultiOn benchmark (standard set) results for 3-object navigation.}
\label{tab:deep_transport}
\centering
\scalebox{0.66}{
\begin{tabular}{c ||c|c|c|c|| c|c|c|c|| c|c|c|c}
\toprule
Model & \multicolumn{4}{c||}{Default} & \multicolumn{4}{c||}{Hard} & \multicolumn{4}{c}{Harder}\\ \midrule
\multicolumn{13}{c}{$\|$ \%Success$\uparrow$ $\vert$ \%Progress$\uparrow$  $\vert$ SPL$\uparrow$ $\vert$ PPL$\uparrow$ $\|$ }\\ \midrule
OracleMap (Occ) & 26     & 45         & 18 & 27 & 3.6   & 17    & 2.3   & 10 & 2     & 12    & 1.3   & 7.7\\
OracleMap (Occ+Obj) &  85      & 91         & 65 & 67 & 46      & 69         & 29 & 38 & 26      & 48         & 16 & 26\\
OracleMap-Waypoints & 80 & 88 & 38 & 39 & 48 & 72 & 18 & 24 & 31 & 59 & 12 & 19\\
NoMap & 39      & 59         & 25 & 31 & 5.2      & 24         & 2.6 & 10 & 2.8      & 16         & 1.3 & 8.0\\
HTP-NearestFrontier & 41 & 54 & 17 & 20 & 22 & 40 & 8.1 & 14 & 15 & 32 & 5.5 & 12\\
HTP-CNN                       & {\bf 55}    & {\bf 68}    & {\bf 26}    & {\bf 30} & {\bf 33}      & {\bf 53}         & {\bf 13} & {\bf 20} & 21      & {\bf 42}         & {\bf 8.6} & {\bf 16}\\
HTP-GCN & 46 & 58 & 19 & 22 & 27 & 47 & 9.4 & 16 & {\bf 22} & 39 & 8.0 & 14\\
\bottomrule
\end{tabular}

\begin{tabular}{c|c|c|c|c}
\toprule
Model & \%Success$\uparrow$ & \%Progress$\uparrow$  & SPL$\uparrow$ & PPL$\uparrow$\\\midrule
OracleMap (Occ) & 16 & 36 & 12 & 27\\
OracleMap (Occ+Obj) & 48 & 62 & 38 & 49\\
NoMap               & 10 & 24 & 4 & 14 \\
FRMQN\cite{frmqn} & 13 & 29 & 24 & 24\\
SMT\cite{fang2019scene} & 9 & 22 & 7 & 18\\
ProjNeuralMap\cite{wani2020multion} & 27 & 46 & 18 & 31\\
ObjRecogMap\cite{wani2020multion} & 22 & 40 & 17 & 30\\
Lyon\cite{marza2021teaching} & 57 & 70 & \textbf{36} & \textbf{45}\\
HTP-CNN (Ours) & 56 & 69 & 30 & 36\\
HTP-GCN (Ours)& \textbf{57} & \textbf{70} & 27 & 33 \\
\bottomrule

\end{tabular}
}

\end{table}

\subsection{Results} \label{ssec:quant_compare}



\paragraph{Standard Long-HOT Task:}  

Table \ref{tab:standard} shows the results of evaluations on Standard Long-HOT task for 1000 test episodes generated using the \emph{default} task level.
All variants of HTP clearly outperform NoMap and ProjNeuralMap on all six  metrics. 
Fig. \ref{fig:example_results} visualizes an episode of HTP-CNN. We show video results of our work in supplementary.

\vspace{0.2cm}
\noindent\emph{Are hierarchies good?} HTP-NearestFrontier already outperforms the non-hierarchical flat baselines by a large margin, showing the importance of our modular hierarchical approach involving separate policies for different task phases, coupled with a topological map. Interestingly, not all hierarchies are good: in particular, OracleMap-Waypoints, which sets waypoint subgoals for a motion planner, performs clearly worse than flat OracleMap (``Occ+Obj''). Note that OracleMap methods have access to ground truth map information and are not directly comparable with HTP, but can be meaningfully compared among themselves. 

\vspace{0.2cm}
\noindent\emph{Does weighted frontier exploration work?} Among HTP variants, both HTP-GCN and HTP-CNN, which use predicted scores for weighted frontier exploration, clearly outperform HTP-NearestFrontier. Between them, GCN and CNN are roughly equivalent in this setting.

\vspace{0.2cm}
\noindent\emph{How important is good agent-centered occupancy and object location information?} On this test set, access to the ground truth occupancy and object maps centered around the agent significantly improves performance, with OracleMap (Occ+Obj) performing the best out of all methods.



\vspace{-0.2cm}
\paragraph{Large Long-HOT Task:} We now evaluate these same trained policies on more challenging settings in large scenes, with more difficult transport task levels. This evaluates generalization and highlights the benefits of effective hierarchy and modularity. Note that scenes used for testing in Large Long-HOT have some overlap with training scenes but not with the same episodes. This is due to limited large scenes available in the disjoint test set. But our observation of better generalization stands since all methods have the same advantage, but others suffer severe drops compared to ours.

Table \ref{tab:deep_transport} (left) shows the results. Flat end-to-end approaches like NoMap and OracleMap deteriorate catastrophically on \textit{hard} and \textit{harder} level tasks. 
NoMap's \%Success drops from 39\% on \emph{default} to 
a mere 5.2\% and 2.8\% on \emph{hard} and \emph{harder} levels. 
HTP  methods degrade more gracefully, achieving up to 33\% and 22\% on \textit{hard} and \textit{harder}. In fact, as task difficulty increases, HTP significantly closes the performance gap to OracleMap-Waypoints despite the oracle  method's access to ground truth map information. We believe this is because our modular approach with weighted frontier exploration leads to better generalization compared to waypoint setting as in OracleMap-Waypoints. Further, OracleMap-Waypoints performs better than OracleMap (Occ+Obj) confirming the benefits of subgoals in long-horizon settings. 
To further study the dependence of HTP's model components on the task performance and its generalization to increasing number of object goals we conduct several ablations, results for which are available in the supplementary. 



Overall, this large effect of small increments in the spatial task scale at \textit{hard} and \textit{harder} levels  (Tab~\ref{tab:dataset}) shows how Long-HOT stress-tests planning, exploration and reasoning over long spatial and temporal horizons. This is different from prior efforts~\cite{wani2020multion} that extend a task by adding new objects, but pre-specify a sequence of single-object sub-tasks.
Our HTP approach, which leverages hierarchical policies and topological maps, is a first step towards addressing these unique challenges induced by Long-HOT. 
Note that, difficulty of Long-HOT is expected to be a lot higher with the end criterion of MultiON, analysis for which will be in the supplementary.

\vspace{-0.2cm}
\paragraph{Results on MultiOn:} Finally, we also evaluate our proposed HTP framework on the MultiOn\cite{wani2020multion} challenge. Table \ref{tab:deep_transport} (right) shows that our method significantly outperforms other baselines from \cite{wani2020multion}. Moreover, its performance is nearly on par or better with the CVPR 2021 Embodied AI workshop challenge winner \cite{marza2021teaching}. Note that the techniques proposed in \cite{marza2021teaching} are complementary to ours. 
\vspace{-0.1cm}
\section{Conclusion}
In this paper, we addressed the problem of long-horizon exploration and planning by introducing a novel Long-HOT benchmark. Further, we proposed a modular hierarchical transport policy (HTP) that builds a topological graph of the scene to perform exploration with the help of weighted frontiers and simplify navigation in long-horizons through a combination of motion planning and RL policy robust to imperfect hand-offs.
Our sub-task policies are connected in novel ways with different levels of hierarchical control requiring different state representations to perform object transport. We show how our approach leads to large improvements in performance on the transport task, it's ability to generalize to harder long-horizon task settings while only training on simpler versions, and also achieve state-of-the-art numbers on MultiON. 


%
%
\bibliographystyle{splncs04}
\bibliography{references}
\end{document}


\pagestyle{headings}
\mainmatter
\def\ECCVSubNumber{5255}  

\title{Long-HOT: A Modular Hierarchical Approach for Long-Horizon Object Transport} 

\titlerunning{Long-HOT}
%
\author{Sriram Narayanan$^1$ \and 
Dinesh Jayaraman$^2$  \and
Manmohan Chandraker$^{1,3}$
}
\authorrunning{Narayanan et al.}
%

\institute{NEC Labs America \and 
University of Pennsylvania \and 
UC San Diego
}
\appendix
\noindent {\Large \bf Appendix}

\begin{abstract}
    In this supplementary, we describe the dataset statistics of LongHOT and provide some more analysis of HTP in terms of its generalizability to increasing number of target objects. Further, we provide some additional implementation details for HTP and its baselines. With the accompanying video we include an overview of our contributions along with qualitative comparison and visualizations. 
\end{abstract}

\section{Further Analysis for Long-HOT and HTP}

\subsection{Ablation Study}
Here, we study the dependence of various components in HTP to the overall task performance and also their generalizability to increasing number of target objects for the transport task. For creating a test episode, we sample additional targets with object colors that were used during the training. 
Note that we only test our methods with increasing number of object goals by training them on a \emph{default} task level with 4 objects. The ablations were conducted using the GCN variant in HTP on 250 test episodes with 15 scenes.

\vspace{-0.3cm}
\paragraph{\bf HTP w/o graph:}
We perform an ablation of HTP without a graph memory for closeness scores. When the high-level action is one of \texttt{\small\{Pickup, Drop\}} action, the Pick-Drop module provides direct control to object-nav policy bypassing the low-level controller. Here, instead of reaching the node with highest closeness score we directly use object-nav policy to move closer towards target objects. 

\begin{figure}[h]
    \centering
    \includegraphics[width=1.0\textwidth, trim={0 0 0 0}, clip]{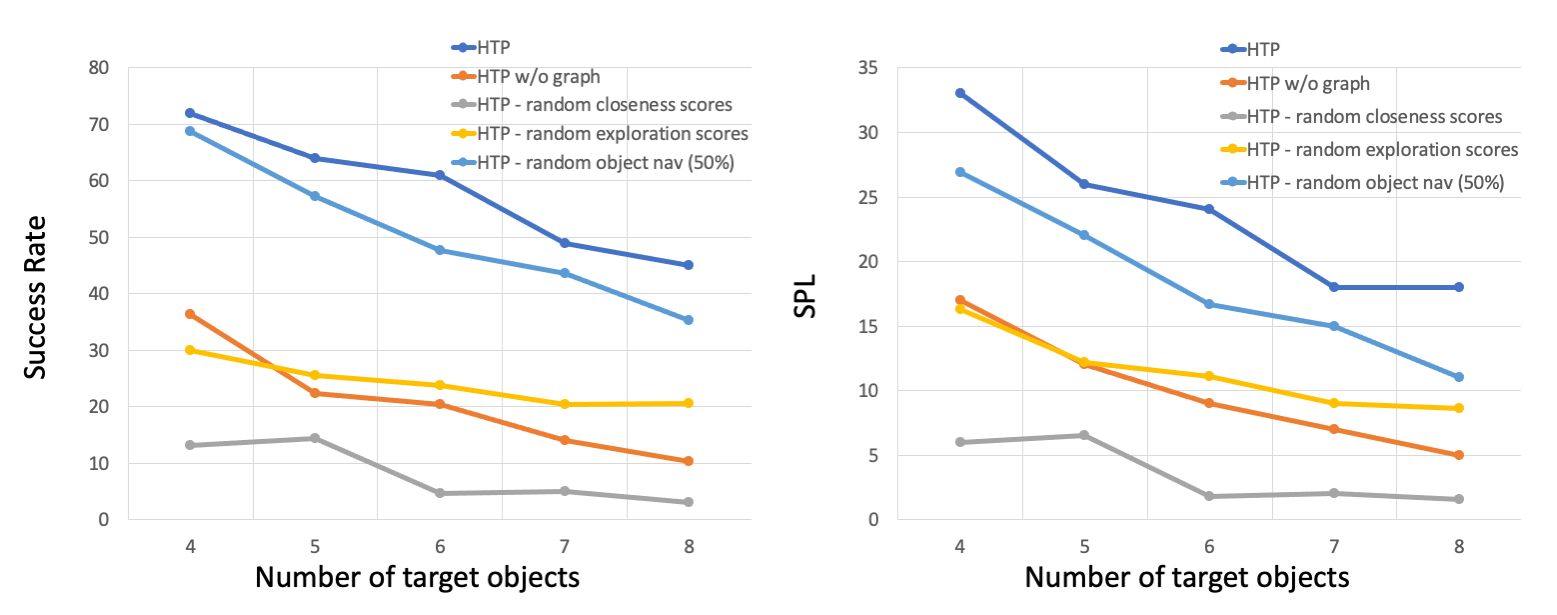}
    \caption{Performance comparison based on success rates and SPL of the proposed HTP with ablations for increasing number of target objects. 
    }
    \label{fig:ablation}
\end{figure}

\vspace{-0.3cm}
\paragraph{\bf HTP - random closeness scores:} To study how the values of closeness scores affect the task performance, we replace closeness values to be random numbers. Intuitively, this should affect agent the most as random closeness values will take agent to a very different node location compared to the one closest to the target object. Further, it also makes agent execute pick-drop operations at unintended locations thereby affecting the task performance. 

\vspace{-0.3cm}
\paragraph{\bf HTP - random exploration scores:} In this ablation, we replace the frontier exploration scores to random values which affects the frontier selection and the exploration strategy. 

\vspace{-0.3cm}
\paragraph{\bf HTP - random object navigation (50\%):} Here we replace the actions from the object navigation policy with random actions 50\% of the time.

Fig. \ref{fig:ablation} reports the success rate and SPL of various HTP variants for increasing number of target objects. As shown, the performance of HTP is considerably higher than the all other ablations indicating the purpose of each of these model components. 
HTP w/o graph provides significant drop in performance compared to HTP due to the long-horizon exploration and navigation required by object-nav for finding target objects. With increasing number of target objects the performance of HTP w/o graph deteriorates even further relatively indicating the importance of topological graph. HTP with random closeness score affect agents the most due to its influence on pick drop operations as incorrect values take agent to a completely different node compared to the closest one and the performance even reduces close to 0\% with increase in number of target objects. 

HTP with random exploration score provides low success rates as incorrect frontier values makes agent switch between frontiers that are farther away making exploration less efficient as the agent travels within the explored regions for significant portion of its time while switching. 
HTP is not affected much even when perturbed with random actions for object navigation 50\% of the time, indicating the robustness of HTP. This could be due to HTP simplifying the navigation process through a combination of motion planning and RL where, we use motion planning for navigation within explored regions and RL policies for navigating towards semantic targets at unknown locations. 

While the performance of all the HTP variants decrease with increase in number of target objects, HTP still provides good enough performance for 8 object transport while only training on transport task involving 4 objects. This also shows HTP's ability to generalize towards increasing number of target objects.



\begin{table*}[]
\centering
\caption{Standard Long-HOT Task with early termination for wrong pickup action similar to MultiOn\cite{wani2020multion}.}
\begin{tabular}{lcccccc}
\toprule
Model                                     & \%Success$\uparrow$ & \%Progress$\uparrow$  & SPL$\uparrow$ & PPL$\uparrow$  & Episode energy$\downarrow$ & \%Picked$\uparrow$ \\
\midrule
OracleMap (Occ) & 0 & 0.08 & 0 & 0.08 & 0.99 & 1.3\\
OracleMap (Occ+Obj) & 6.4 & 16.5 & 5.92 & 15.8 & 0.83 & 28       \\
ProjNeuralMap\cite{wani2020multion} & 0.1 & 1.6 & 0.1 & 1.6 & 0.98 & 6.2\\
HTP - GCN & 54 & 67 & 26 & 32 & 0.37 & 75\\
\bottomrule
\end{tabular}
\label{tab:standard_wterminate}
\end{table*}
\subsection{Episode termination for wrong pickup}
Table \ref{tab:standard_wterminate} compares the performance of HTP and baseline methods with early termination criteria for a wrong pickup action. An episode is terminated if there are no objects within agent's vicinity when a pickup action is called. The numbers indicate Long-HOT's difficulty to be a lot higher with an end criterion similar to MultiOn\cite{wani2020multion}. We relax those hard constraints as training becomes more difficult with sparse rewards for long-horizon tasks and rather focus on task completion for our agents. While the performance of RL methods drop significantly in Table \ref{tab:standard_wterminate}, the proposed HTP is nearly unaffected. It can be attributed to the modular hierarchical design of HTP where pick-drop actions are executed only when the object closeness score is higher than a threshold.


\subsection{Long-HOT episode statistics}
\begin{figure}[t]
    \centering
    \begin{tabular}{cccc}
         \rotatebox[origin=l]{90}{\makebox[1in]{Standard Long-HOT}} & 
         \includegraphics[width=0.3\textwidth]{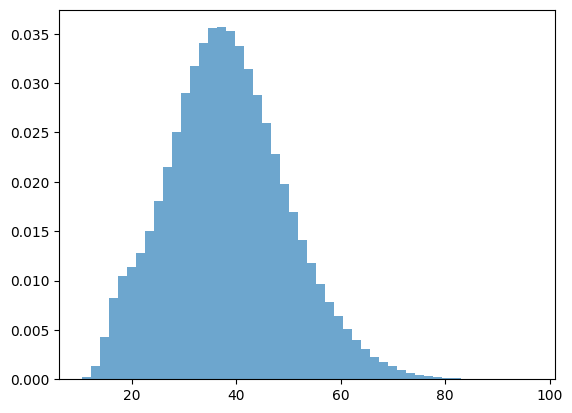} &
         \includegraphics[width=0.3\textwidth]{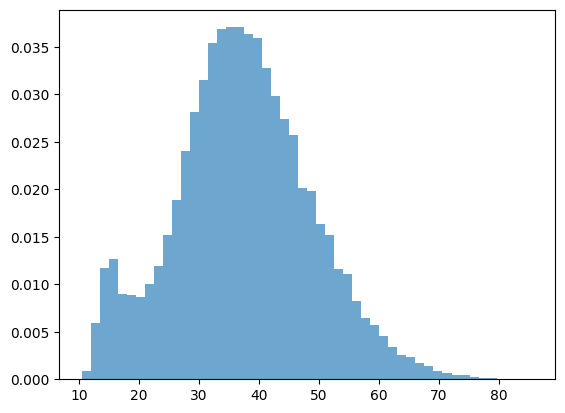} & 
         \includegraphics[width=0.3\textwidth]{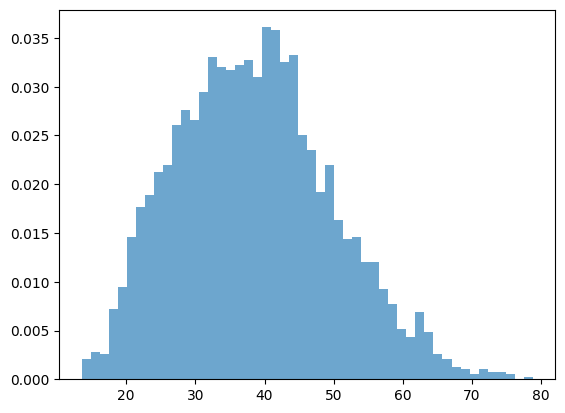} \\ [0.3cm]
         
         \rotatebox[origin=l]{90}{\makebox[1in]{\small Large Long-HOT}} & 
         \includegraphics[width=0.3\textwidth]{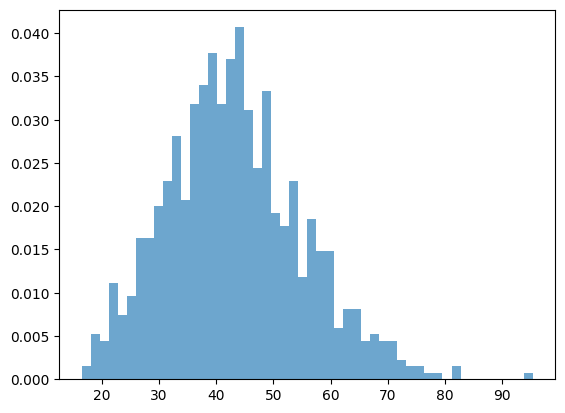} & 
         \includegraphics[width=0.3\textwidth]{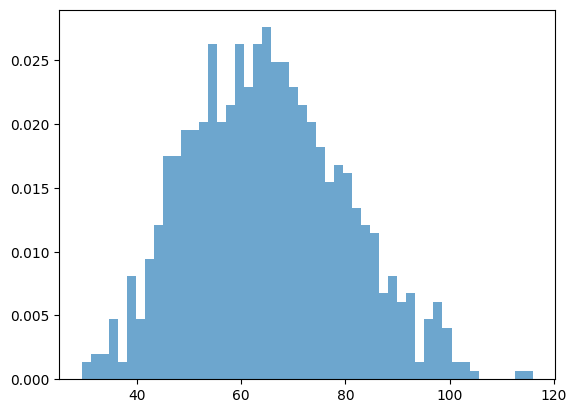} &
         \includegraphics[width=0.3\textwidth]{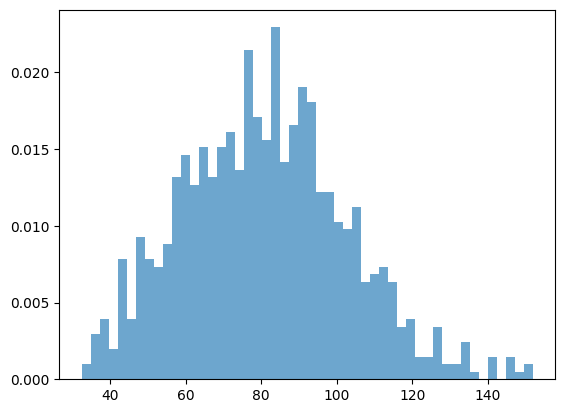}
         
    \end{tabular}
    \caption{Shows the statistics of the episodes used in Long-HOT. The x-axis represents geodesic distance of a reference trajectory in meters and y-axis shows fraction of total episodes in the corresponding histogram bin. 
    (a) represents statistics of the episodes used to train all methods. (c, d, e and f) show test data statistics used in standard and large Long-HOT settings. Geodesic distance of the reference trajectories in Large Long-HOT, indicates increasing task complexity of test scenarios where agent's require deeper exploration and long-horizon planning to complete the task.
    }
    \label{fig:dataset_statistics}
    
\end{figure}

Fig. \ref{fig:dataset_statistics} shows a histogram of geodesic distances with fraction of total episodes in the corresponding histogram bin for various datasets used in Long-HOT experiments. Note the range of reference path length increases with different task configurations in Large Long-HOT indicating a increasing task complexity.

\vspace{-0.4cm}
\subsection{Discussion and limitations}
The accompanying video shows some failure cases of HTP which includes situations where an agent is unable to move around an obstacle or has small frontier regions which are ignored by the exploration module and situations where the closeness scores for some object visible only from some particular node direction is overwritten by values obtained when the perspective image does not contain the object viewed from a different location localized to the same node direction.

The work assumes noiseless odometry and depth for task completion, but earlier works like \cite{chaplot2020object} have shown that semantic mapping and navigation work well in the real world even with noisy pose and depth. Future works can relax these assumptions to build methods that work more robustly with different forms of noisy inputs.

\section{Further Details on Implementation, Training and Metrics}

In this section, we provide additional implementation details of baseline architectures and the proposed HTP.

\subsection{No Map baseline}

We adapt an architecture similar to \cite{wani2020multion} for the No Map policy. 

\vspace{-0.4cm}
\paragraph{Inputs and Outputs:} No Map takes an RGBD image of size $256\times256$ along with the hand state $O_h$ (size $5\times1$), goal state $O_g$ (size $4\times1$) and previous action as inputs to the policy. It then predicts one of \texttt{\{\small FORWARD, TURN LEFT, TURN RIGHT, PICKUP, DROP\}} actions at every timestep.

\vspace{-0.4cm}
\paragraph{Architecture:} The RGBD image is passed through a sequence of three convolutional layers + ReLU\cite{relu} and a linear layer + ReLU that transforms the input into a feature vector of length $512$. The convolutional layers consist of kernels with size $\{8,4,3\}$, strides $\{4,2,1\}$ and output channels $\{32,64,32\}$ respectively. The hand state $O_h$ and goal state $O_g$ are passed through dense layers to get respective feature vectors (dim $32$). The previous action is embedded through an embedding layer of length $32$. Finally, image features, hand-goal features, and previous action embedding are concatenated and passed through a recurrent unit to output features that are used to predict actions and the approximate value function.


\vspace{-0.4cm}
\paragraph{Rewards:} The following rewards $r_t$ is provided at every timestep $t$ to train the No Map agent:
\begin{equation}
    \begin{split}
        r_t = & \mathbb{I}_{success} \cdot r_{success} + \mathbb{I}_{pick} \cdot r_{pick} + r_{d2o} +r_{d2g} \\ & + \sum_o \mathbb{I}^o_{goal} \cdot r_{goal} + r_{collision} +  r_{fpd} + r_{slack}
    \end{split}
\end{equation}
where, $\mathbb{I}_{success}, \mathbb{I}_{pick}, \mathbb{I}^o_{goal}$ are functions that indicate successful completion of the episode, any target object picked for the first time and target objects that are transported to the goal respectively. $r_{success}, r_{pick}, r_{goal}$ are the rewards associated with $\mathbb{I}_{success}, \mathbb{I}_{pick}, \mathbb{I}^o_{goal}$. $r_{d2o} = (d^o_{t-1} - d^o_t)$ is the decrease in geodesic distance from the agent's position to the closest object. $r_{d2g} = \max_o (d^o_{t-1} - d^o_t)$ is the maximum decrease in geodesic distance of target objects with the goal. $r_{collision}$ is the collision penalty for agents and $r_{slack}$ is the slack reward for every timestep the agent delays in completing the episode.


\vspace{-0.4cm}
\paragraph{Training:} The policy is trained using proximal policy optimization (PPO) \cite{ppo} technique, for approximately 40M iterations using 24 simulator instances. The hyper-parameters used are similar to \cite{wani2020multion}.

\subsection{OracleMap Baselines}
We first describe the OracleMap (Occ) and OracleMap (Occ+Obj) baselines and then provide details on OracleMap-Waypoints policy.

\subsubsection{OracleMap (Occ / Occ+Obj)}
The policy architecture for OracleMap (Occ / Occ + Obj) is similar to No Map agent with an additional map input that covers an area of $10m\times10m$. First top view map embeddings (dim. 16) are generated and then passed through a map encoder. The encoder consists of convolutions with kernels $\{4,3,2\}$, stride $\{3,1,1\}$ and output channels $\{32,64,32\}$. The map encoder produces a feature vector of length $256$ and is concatenated as one of the inputs to the recurrent unit. The output action space and rewards used to train OracleMap (Occ / Occ + Obj) is similar to the No Map baseline.

\subsubsection{OracleMap-Waypoints}
The inputs to the baseline are OracleMap (Occ+Obj), hand state $O_h$ and goal state $O_g$. It then predicts waypoints to be reached as $(x,y)$ locations on the map. The prediction is discretized into $M=100$ bins within a $5m\times5m$ range centered on the agent. The agent then selects a bin as one of its action and uses A*\cite{astar} to reach its location. The predicted subgoal is also associated with one of \texttt{\small \{Pickup, Drop\}} high-level actions. The action space in Waypoints policy contains $M\times2$ in total. Once the agent reaches the predicted subgoal, the corresponding high level action \texttt{\small \{Pickup, Drop\}} is executed. The subgoals are updated for every $t_k$ steps irrespective of agent reaching previously assigned subgoal. Agent's prediction range is kept higher than maximum traversable distance in $t_k$ steps for agents to provide subgoals that avoid taking pickup or drop actions within $t_k$ timesteps.  The policy architecture and rewards used are similar to OracleMap (Occ+Obj) baselines.



\subsection{Hierarchical Transport Policy}

\paragraph{High-level Controller:} 
The controller takes hand state $O_h$, goal state $O_g$ and closeness scores $\mathcal{F}_O$ of objects with respect to the nodes as inputs. The high-level action is assigned based on the following conditions, if the agent already holds maximum number of objects in its hand and goal object was discovered (closeness score of goal $> F_O^{th}$ in one of the nodes) a \texttt{\{Drop\}} action is executed else the agent executes \texttt{\{Explore\}} to find the goal object. If the agent has capacity to carry more objects, and some closeness scores for objects are greater than $F_O^{th}$ in the nodes, a \texttt{\{Pickup[Object]\}} action is executed for objects that are not either held by the agent or transported already. The agent executes \texttt{\{Explore\}} if none of the objects satisfy the closeness score criteria. \texttt{\{Pickup[Container]\}} action is only executed when the container is discovered and the agent does not hold any object in its hand.

\paragraph{Pre-trained Encoder $\mathcal{F}_A$ :}
Latent features from a pre-trained auto-encoder is used as node features in HTP. It consists of a ResNet-18\cite{resnet18} style encoder-decoder architecture with latent features of $32$ dimensions. The auto-encoder is trained with a weighted MSE loss that weighs pixels of target objects with a weight $\lambda=2.0$.


\subsection{Reference Trajectory Calculation}
A reference demonstration in the transport task first picks up the container and then picks up consecutive closest objects from its previous location to finally drop them at goal. The sum of geodesic distances in executing this reference trajectory in an episode from agent's starting location is used as the reference path length $G_{ref}$.



%
%
\bibliographystyle{splncs04}
\bibliography{references}